\def\year{2019}\relax
\documentclass[letterpaper]{article} 
\usepackage{aaai19}  
\usepackage{times}  
\usepackage{helvet}  
\usepackage{courier}  
\usepackage{url}  
\usepackage{graphicx}  
\graphicspath {img/}
\usepackage{amssymb,amsmath,mathtools}
\usepackage[ruled]{algorithm2e}
\usepackage{subfig}
\usepackage{color}
\usepackage{bm}
\newtheorem{theorem}{Theorem}

\newcommand{\citet}[1]{\citeauthor{#1} ̃\shortcite{#1}}
\newcommand{\citep}{\cite}
\newcommand{\citealt}[1]{\citeauthor{#1} ̃\citeyear{#1}}

\begin{document}
%
\title{ACE: An Actor Ensemble Algorithm for Continuous Control with Tree Search}
\author{Shangtong Zhang$^1$~\thanks{Work done during an internship at Huawei}, Hao Chen$^2$, Hengshuai Yao$^2$\\
$^1$ Department of Computing Science, University of Alberta\\
$^2$ Reinforcement Learning for Autonomous Driving Lab, Noah's Ark Lab, Huawei\\
shangtong.zhang@ualberta.ca, \{hao.chen, hengshuai.yao\}@huawei.com\\
}
\maketitle
\begin{abstract}
In this paper, we propose an actor ensemble algorithm, named ACE, for continuous control with a deterministic policy in reinforcement learning. In ACE, we use actor ensemble (i.e., multiple actors) to search the global maxima of the critic. Besides the ensemble perspective, we also formulate ACE in the option framework by extending the option-critic architecture with deterministic intra-option policies, revealing a relationship between ensemble and options. Furthermore, we perform a look-ahead tree search with those actors and a learned value prediction model, resulting in a refined value estimation. We demonstrate a significant performance boost of ACE over DDPG and its variants in challenging physical robot simulators. 
\end{abstract}

\section{Introduction}

\noindent In this paper, we propose an actor ensemble algorithm, named ACE, for continuous control in reinforcement learning (RL). In continuous control, a deterministic policy (\citealt{silver2014deterministic}) is a recent approach, which is a mapping from state to action. In contrast, a stochastic policy is a mapping from state to a probability distribution over the actions. 

Recently, neural networks has achieved great success as function approximators in various challenging domains (\citealt{tesauro1995temporal}; \citealt{mnih2015human}; \citealt{silver2016mastering}). A deterministic policy parameterized by a neural network is usually trained via gradient ascent to maximize the critic, which is a state-action value function parameterized by a neural network (\citealt{silver2014deterministic}; \citealt{lillicrap2015continuous}; \citealt{barth2018distributed}). However, gradient ascent can be easily trapped by local maxima during the search for the global maxima. We utilize the ensemble technique to mitigate this issue. We train multiple actors (i.e., deterministic policies) in parallel, and each actor has a different initialization. In this way, each actor is in charge of maximizing the state-action value function in a local area. Different actors may be trapped in different local maxima. By considering the maximum state-action value of the actions proposed by all the actors, we are more likely to find the global maxima of the state-action value function than a single actor. 

ACE fits in with the option framework (\citealt{sutton1999between}). First, each option has its intra-option policy, which maximizes the return in a certain area of the state space. Similarly, an actor in ACE maximizes the critic in a certain area of the domain of the critic. It may be difficult for a single actor to maximize the critic in the whole domain due to the complexity of the manifold of the critic. However, in contrast, the job for action search is easier if we ask an actor to find the best action in a local neighborhood of the action dimension. Second, we chain the outputs of all the actors to the critic, enabling a selection over the locally optimal action values. In this way, the critic works similar to the policy over options in the option framework. We quantify this similarity between ensemble and options by extending the option-critic architecture (OCA, \citealt{bacon2017option}) with deterministic intra-option policies. Particularly, we provide the Deterministic Intra-option Policy Gradient theorem, based on which we show the actor ensemble in ACE is a special case of the general option-critic framework.

To make the state-action value function more accurate, which is essential in the actor selection, we perform a look-ahead tree search with the multiple actors. The look-ahead tree search has achieved great success in various discrete action problems (\citealt{knuth1975analysis}; \citealt{browne2012survey}; \citealt{silver2016mastering}; \citealt{oh2017value}; \citealt{farquhar2018treeqn}). Recently, look-ahead tree search was extended to continuous-action problems. For example, \citet{mansley2011sample} combined planning with adaptive discretization of a continuous action space, resulting in a performance boost in continuous bandit problems. \citet{nitti2015planning} utilized probability programming in planning in continuous action space. \citet{yee2016monte} used kernel regression to generalize the values between explored actions and unexplored actions, resulting in a new Monte Carlo tree search algorithm. However, to our best knowledge, a general tree search algorithm for continuous-action problems is a gap. In ACE, we use the multiple actors as meta-actions to perform a tree search with the help of a learned value prediction model (\citealt{oh2017value}; \citealt{farquhar2018treeqn}).

We demonstrate the superiority of ACE over DDPG and its variants empirically in Roboschool~\footnote{https://github.com/openai/roboschool}, a challenging physical robot environment. 

In the rest of this paper, we first present some preliminaries of RL. Then we detail ACE and show some empirical results, after which we discuss some related work, followed by closing remarks.

\section{Preliminaries} 

We consider a Markov Decision Process (MDP) which consists of a state space $\mathcal{S}$, an action space $\mathcal{A}$, a reward function $r: \mathcal{S} \times \mathcal{A} \rightarrow \mathbb{R}$, a transition kernel $p: \mathcal{S} \times \mathcal{A} \times \mathcal{S} \rightarrow [0, 1]$, and a discount factor $\gamma \in [0, 1]$. We use $\pi: \mathcal{S} \times \mathcal{A} \rightarrow [0, 1]$ to denote a stochastic policy. At each time step $t$, an agent is at state $s_t$ and selects an action $a_t \sim \pi(\cdot|s_t)$. Then the environment gives the reward $r_{t+1}$ and leads the agent to the next state $s_{t+1}$ according $p(\cdot|s_t, a_t)$. We use $q^\pi$ to denote the state action value of a policy $\pi$, i.e., $q^\pi (s, a) \doteq \mathbb{E}_\pi[\sum_{t=0}^\infty \gamma^t r_{t+1} | s_0 = s, a_0 = a]$. In an RL problem, we are usually interested in finding an optimal policy $\pi^*$ s.t. $q^{\pi^*}(s, a) \geq q^\pi(s, a)$ for $\forall (\pi, s, a)$. All the optimal policies share the same optimal state action value function $q^*$, which is the unique fixed point of the Bellman optimality operator $\mathcal{T}$,
\begin{align}
\label{eq:bellman}
\mathcal{T}Q(s, a) \doteq r(s, a) + \gamma \mathbb{E}_{s^\prime \sim p(\cdot|s, a)}[\max_{a^\prime} Q(s^\prime, a^\prime)]
\end{align}
where we use $Q$ to indicate an estimation of $q^*$. With tabular representation, Q-learning (\citealt{watkins1992q}) is a commonly used method to find this fixed point. The per step update is 
\begin{align}
\label{eq:q}
\Delta Q(s_t, a_t) \propto r_{t+1} + \gamma \max_{a} Q(s_{t+1}, a) - Q(s_t, a_t)
\end{align}
Recently, \citet{mnih2015human} used a deep convolutional neural network $\theta$ to parameterize an estimation $Q$, resulting in the Deep-Q-Network (DQN). At each time step $t$, DQN samples the transition $(s_t, a_t, r_{t+1}, s_{t+1})$ from a replay buffer (\citealt{lin1992self}) and performs stochastic gradient descent to minimize the loss 
\begin{align*} 
\frac{1}{2}\big(r_{t+1} + \gamma \max_aQ_{\theta^-}(s_{t+1}, a) - Q_\theta(s_t, a_t)\big)^2
\end{align*}
where $\theta^-$ is a target network (\citealt{mnih2015human}), which is a copy of $\theta$ and is synchronized with $\theta$ periodically.

\subsection{Continuous Action Control}
In continuous control problems, it is not straightforward to apply Q-learning. The basic idea of the deterministic policy algorithms (\citealt{silver2014deterministic}) is to use a deterministic policy $\mu: \mathcal{S} \rightarrow \mathcal{A}$ to approximate the greedy action $\arg\max_a Q(s, a)$. The deterministic policy is trained via gradient ascent according to chain rule. The gradient per step is
\begin{align}
\label{eq:du}
\nabla_{\theta^\mu} Q(s_t, \mu(s_t)) = \nabla_a Q(s, a)|_{s=s_t, a=\mu(s_t)}\nabla_{\theta^\mu}\mu(s)|_{s=s_t}
\end{align}
where we assume $\mu$ is parameterized by $\theta^\mu$. With this actor $\mu$, we are able to update the state action value function $Q$ as usual. To be more specific, assuming $Q$ is parameterized by $\theta^Q$, we perform gradient decent to minimize the following loss 
\begin{align*}  
\frac{1}{2}\big(r_{t+1} + \gamma Q(s_{t+1}, \mu(s_{t+1})) - Q(s_t, a_t)\big)^2
\end{align*}
Recently, \citet{lillicrap2015continuous} used neural networks to parameterize $\mu$ and $Q$, resulting in the Deep Deterministic Policy Gradient (DDPG) algorithm. DDPG is an off-policy control algorithm with experience replay and a target network. 

In the on-policy setting, the gradient in Equation~\eqref{eq:du} guarantees policy improvement thanks to the Deterministic Policy Gradient theorem (\citealt{silver2014deterministic}). In off-policy setting, the policy improvement of this gradient is based on Off-policy Policy Gradient theorem (OPG, \citealt{degris2012off}). However, Errata of \citet{degris2012off} shows that OPG only holds for tabular function representation and certain linear function approximation. So far no policy improvement is guaranteed for DDPG. However, DDPG and its variants has gained great success in various challenging domains (\citealt{lillicrap2015continuous}; \citealt{barth2018distributed}; \citealt{tassa2018deepmind}). This success may be attributed to the gradient ascent via chain rule in Equation~\eqref{eq:du}. 

\subsection{Option}
An option $\omega$ is a triple, $(\mathcal{I}_\omega, \pi_\omega, \beta_\omega)$, and we use $\Omega$ to indicate the option set. We use $\mathcal{I}_\omega \subseteq \mathcal{S}$ to denote the initiation set of $\omega$, indicating where the option $\omega$ can be initiated. In this paper, we consider $\mathcal{I}_\omega \equiv \mathcal{S}$ for $\forall \omega$, meaning that each option can be initiated at all states. We use $\pi_\omega: \mathcal{S} \times \mathcal{A} \rightarrow [0, 1]$ to denote the intra-option policy of $\omega$. Once the option $\omega$ is committed, the action selection is based on $\pi_\omega$. We use $\beta_\omega: \mathcal{S} \rightarrow [0, 1]$ to denote the termination function of $\omega$. At each time step $t$, the agent terminates the previous option $\omega_{t-1}$ with probability $\beta_{\omega_{t-1}}(s_t)$. In this paper, we consider the \textit{call-and-return} option execution model (\citealt{sutton1999between}), where an agent executes an option until the option terminates.

An MDP augmented with options forms a Semi-MDP (\citealt{sutton1999between}). We use $\pi_\Omega: \mathcal{S} \times \Omega \rightarrow [0, 1]$ to denote the policy over options. We use $Q_\Omega: \mathcal{S} \times \Omega \rightarrow \mathbb{R}$ to denote the option-value function and $V_\Omega: \mathcal{S} \rightarrow \mathbb{R}$ to denote the value function of $\pi_\Omega$. Furthermore, we use $U: \Omega \times \mathcal{S} \rightarrow \mathbb{R}$ to denote the option value upon arrival at the state option pair $(\omega, s^\prime)$ and $Q_U: \mathcal{S} \times \Omega \times \mathcal{A} \rightarrow \mathbb{R}$ to denote the value of executing an action $a$ in the context of a state-option pair $(s, \omega)$. They are related as
\begin{align}
Q_\Omega(s, \omega) &= \sum_a \pi_\omega(a | s) Q_U(s, \omega, a) \nonumber \\
\label{eq:def_q_U}
Q_U(s, \omega, a) &= r(s, a) + \gamma \mathbb{E}_{s^\prime \sim p(\cdot|s, a)}U(\omega, s^\prime) \\
\label{eq:def_U}
U(\omega, s^\prime) &= \big(1 - \beta_\omega(s^\prime)\big)Q_\Omega(s^\prime, \omega) + \beta_\omega(s^\prime)V_\Omega(s^\prime) \\
\label{eq:def_V}
V_\Omega(s) &= \sum_\omega \pi_\Omega(\omega | s) Q_\Omega(s, \omega) 
\end{align}
\citet{bacon2017option} proposed a policy gradient method, the Option-Critic Architecture (OCA), to learn stochastic intra-option policies $\{\pi_\omega\}$ (parameterized by $\theta^\pi$) and termination functions $\{\beta_\omega\}$ (parameterized by $\theta^\beta$). The objective is to maximize the expected discounted return per episode, i.e., $Q_\Omega(s_0, \omega_0)$. Based on their Intro-option Policy Gradient Theorem and Termination Gradient Theorem, the per step updates for $\theta^\pi$ and $\theta^\beta$ are
\begin{align*}
\Delta\theta^\pi &\propto \nabla_{\theta^\pi }\log\pi_{\omega_t}(a_t | s_t)Q_U(s_t, \omega_t, a_t) \\
\Delta\theta^\beta &\propto -\nabla_{\theta^\beta}\beta_{\omega_{t-1}}(s_t)\big(Q_\Omega(s_t, \omega_{t-1}) - V_\Omega(s_t)\big)
\end{align*}
And at each time step $t$, OCA takes one gradient descent step minimizing
\begin{align} 
\label{eq:QU}
\frac{1}{2}\big(g_t - Q_U(s_t, \omega_t, a_t)\big)^2
\end{align}
to update the critic $Q_U$, where
\begin{align*}
g_t \doteq r_{t+1} &+ \gamma \big(1 - \beta_{\omega_t}(s_{t+1})\big)Q_\Omega(s_{t+1}, \omega_t) \\ &+ \gamma \beta_{\omega_t}(s_{t+1}) \max_{\omega^\prime}Q_\Omega(s_{t+1}, \omega^\prime)
\end{align*}
The update target $g_t$ is also used in Intro-option Q-learning (\citealt{sutton1999between}).


\subsection{Model-based RL}
In RL, a transition model usually takes as inputs a state-action pair and generates the immediate reward and the next state. A transition model reflects the dynamics of an environment and can either be given or learned. A transition model can be used to generate imaginary transitions for training the agent to increase data efficiency (\citealt{sutton1990integrated}; \citealt{yao2009multi}; \citealt{gu2016continuous}). A transition model can also be used to reason about the future. When making a decision, an agent can perform a look-ahead tree search (\citealt{knuth1975analysis}; \citealt{browne2012survey}) with a transition model to maximize the possible rewards. A look-ahead tree search can be performed with either a perfect model (\citealt{coulom2006efficient}; \citealt{sturtevant2008analysis}; \citealt{silver2016mastering}) or a learned model (\citealt{weber2017imagination}).

A latent state is an abstraction of the original state. A latent state is also referred to as an abstract state (\citealt{oh2017value}) or an encoded state (\citealt{farquhar2018treeqn}). A latent state can be used as the input of a transition model. Correspondingly, the transition model then predicts the next latent state, instead of the original next state. A latent state is particularly useful for high dimensional state space (e.g., images), where a latent state is usually a low dimensional vector. 

Recently, some works demonstrated that learning a value prediction model instead of a transition model is effective for a look-ahead tree search. For example, VPN (\citealt{oh2017value}) predicted the value of the next latent state, instead of the next latent state itself. Although this value prediction was explicitly done in two phases (predicting the next latent state and then predicting its value), the loss of predicting the next latent state was not used in training. Instead, only the loss of the value prediction for the next latent state was used. \citet{oh2017value} showed this value prediction model is particularly helpful for non-deterministic environments. TreeQN (\citealt{farquhar2018treeqn}) adopted a similar idea, where only the outcome value of a look-ahead tree search is grounded in the loss. \citet{farquhar2018treeqn} showed that grounding the predicted next latent state did not bring in a performance boost. Although a value prediction model predicts much fewer information than a transition model, VPN and TreeQN demonstrated improved performance over baselines in challenging domains. This value prediction model is particularly helpful for a look-ahead tree search in non-deterministic environments. In ACE, we followed these works and built a value prediction model similar to TreeQN. 

\section{The Actor Ensemble Algorithm}
As discussed earlier, it is important for the actor to maximize the critic in DDPG. \citet{lillicrap2015continuous} trained the actor via gradient ascent. However, gradient ascent can easily be trapped by local maxima or saddle points.

To mitigate this issue, we propose an ensemble approach. In our work, we have $N$ actors $\{\mu_1, \dots, \mu_N\}$. At each time step $t$, ACE selects the action over the proposals by all the actors,
\begin{align}
\label{eq:vote}
a_t \doteq \text{argmax}_{a \in \{\mu_i(s_t)\}_{i=1, \dots, N}}Q(s_t, a) 
\end{align}
Our actors are trained in parallel. Assuming those actors are parameterized by $\theta$, each actor $\mu_i$ is trained such that given a state $s$ the action $a = \mu_i(s)$ maximizes $Q(s, a)$. We adopted similar gradient ascent as DDPG. The gradient at time step $t$ is 
\begin{align}
\label{eq:actor-1}
\nabla_{\theta} Q(s_t, \mu_i(s_t)) = \nabla_a Q(s, a)|_{s=s_t, a=\mu_i(s_t)}\nabla_{\theta}\mu_i(s)|_{s=s_t} 
\end{align}
for all $i \in \{1, \dots, N\}$. ACE initializes each actor independently. So that the actors are likely to cover different local maxima of $Q(s, a)$. By considering the maximum action value of the proposed actions, the action in Equation~\eqref{eq:vote} is more likely to find the global maxima of the critic $Q$. To train the critic $Q$, ACE takes one gradient descent step at each time $t$ minimizing
\begin{align} 
\label{eq:critic-1}
\frac{1}{2}\big(r_{t+1} + \gamma \text{max}_{i\in \{1, \dots, N\}} Q(s_{t+1}, \mu_i(s_{t+1})) - Q(s_t, a_t)\big)^2
\end{align}
Note our actors are not independent. They reinforce each other by influencing the critic update, which in turn gives them better policy gradient.

\subsection{An Option Perspective}
Intuitively, each actor in ACE is similar to an option. To quantify the relationship between ACE and the option framework, we first extend OCA with deterministic intra-option policies, referred to as OCAD. For each option $\omega$, we use $\mu_\omega: \mathcal{S} \rightarrow \mathcal{A}$ to denote its intra-option policy, which is a deterministic policy. The intro-option policies $\{\mu_\omega\}$ are parameterized by $\theta$. The termination functions $\{\beta_\omega\}$ are parameterized by $\nu$. We have

\begin{theorem}[Deterministic Intra-option Policy Gradient]
\label{the:DIPG}
Given a set of Markov options with deterministic intra-option policies, the gradient of the expected discounted return objective w.r.t. $\theta$ is:
\begin{align*}
\sum_{s, \omega}\rho_\Omega(s, \omega | s_0, \omega_0)\nabla_a Q_U(s, \omega, a)|_{a=\mu_\omega(s)}\nabla_\theta \mu_\omega(s)
\end{align*}
where $\rho_\Omega(s, \omega | s_0, \omega_0) = \sum_{k=0}^{\infty} P^{(k)}_\gamma(s, \omega | s_0, \omega_0)$ is the limiting state-option pair distribution. Here $P^{(k)}_\gamma$ represents the $\gamma$-discounted probability of transitioning to $(s, \omega)$ from $(s_0, \omega_0)$ in $k$ steps.
\end{theorem}

\begin{theorem}[Termination Policy Gradient]
\label{the:TPG}
Given a set of Markov options with deterministic intra-option policies, the gradient of the expected discounted return objective w.r.t. to $\nu$ is:
\begin{align*}
\sum_{\omega, s^\prime}\rho_\Omega(s^\prime, \omega | s_1, \omega_0)\nabla_\nu \beta_\omega(s^\prime)(V_\Omega(s^\prime) - Q_\Omega(s^\prime, \omega))
\end{align*}
\end{theorem}

The proof of Theorem~\ref{the:DIPG} follows a similar scheme of \citet{sutton2000policy}, \citet{silver2014deterministic}, and \citet{bacon2017option}. The proof of Theorem~\ref{the:TPG} follows a similar scheme of \citet{bacon2017option}. The conditions and proofs of both theorems are detailed in Supplementary Material. The critic update of OCAD remains the same as OCA (Equation~\ref{eq:QU}).

We now show that the actor ensemble in ACE is a special setting of OCAD. The gradient update of the actors in ACE (Equation~\ref{eq:actor-1}) can be justified via Theorem~\ref{the:DIPG}. The critic update in ACE (Equation~\ref{eq:critic-1}) is equivalent to the critic update in OCAD (Equation~\ref{eq:critic-2}). We first consider a special setting, where $\beta_\omega(s) \equiv 1$ for $\forall (\omega, s)$, which means each option terminates at every time step. In this setting, the value of $Q_U(s, \omega, a)$ does not depend on $\omega$ (Equations \ref{eq:def_q_U} and \ref{eq:def_U}). Based on this observation, we rewrite $Q_U(s, \omega, a)$ as $Q(s, a)$. We have 
\begin{align} 
\label{eq:three_q}
Q_\Omega(s, \omega) = Q_U(s, \omega, \mu_\omega(s)) = Q(s, \mu_\omega(s))
\end{align}
With the three $Q$s being the same, we rewrite the intra-policy gradient update in OCAD according to Theorem~\ref{the:DIPG} as 
\begin{align}
\label{eq:actor-2}
\nabla_a Q(s, a)|_{s_t=s, a=\mu_{\omega_t}(s_t)}\nabla_\theta \mu_{\omega_t}(s)|_{s_t=s}
\end{align}
And we rewrite the critic update in OCAD (Equation~\ref{eq:QU}) as
\begin{align}
\label{eq:critic-2}
\frac{1}{2}\big(r_{t+1} + \gamma \max_{\omega^\prime}Q(s_{t+1}, \mu_{\omega^\prime}(s_{t+1})) - Q(s_t, a_t)\big)^2
\end{align}
Now the actor-critic updates in OCAD (Equations \ref{eq:actor-2} and \ref{eq:critic-2}) recover the actor-critic updates in ACE (Equations \ref{eq:actor-1} and \ref{eq:critic-1}), revealing the relationship between the ensemble in ACE and OCAD. 

Note that in the intra-option policy update of OCAD (Equation~\ref{eq:actor-2}) only one intra-option policy is updated at each time step, while in the actor ensemble update (Equation~\ref{eq:actor-1}) all actors are updated. Based on the intro-option policy update of OCAD, we propose a variant of ACE, named Alternative ACE (ACE-Alt), where only the selected actor is updated at each time step. In practice, we add exploration noise for each action $a_t$ and use experience replay to stabilize the training of the neural network function approximator like DDPG, resulting in off-policy learning. 

\subsection{Model-based Enhancements}
To refine the state-action value function estimation, which is essential for actor selection, we utilize a look-ahead tree search method with a learned value prediction model similar to TreeQN. TreeQN was developed for discrete action space. We extend TreeQN to continuous control problems via the actor ensemble by searching over the actions proposed by the actors. 

Formally speaking, we first define the following learnable functions:
\begin{itemize}
\item $f_{enc}: \mathcal{S} \rightarrow \mathbb{R}^n$, an encoding function that transforms a state into an $n$-dimensional latent state, parameterized by $\theta^{enc}$
\item $f_{rew}: \mathbb{R}^n \times \mathcal{A} \rightarrow \mathbb{R}$, a reward prediction function that predicts the immediate reward given a latent state and an action, parameterized by $\theta^{rew}$
\item $f_{trans}: \mathbb{R}^n \times \mathcal{A} \rightarrow \mathbb{R}^n$, a transition function that predicts the next latent state given a latent state and an action, parameterized by $\theta^{trans}$
\item $f_q: \mathbb{R}^n \times \mathcal{A} \rightarrow \mathbb{R}$, a value prediction function that computes the value for a pair of a latent state and an action, parameterized by $\theta^{q}$
\item $f_{\mu_i}: \mathbb{R}^n \rightarrow \mathcal{A}$: an actor that computes an action given a latent state, parameterized by $\theta^{\mu_i}$, for each $i \in \{1, \dots, N\}$   
\end{itemize}
We use $\theta^Q$ to denote $\{\theta^{enc}, \theta^{rew}, \theta^{trans}, \theta^{q}\}$ and $\theta^\mu$ to denote $\{\theta_{enc}, \theta^{\mu_1}, \dots, \theta^{\mu_N}\}$. Note the encoding function is shared in our implementation.

\begin{figure*}[ht]
\centering
\includegraphics[width=0.8\textwidth]{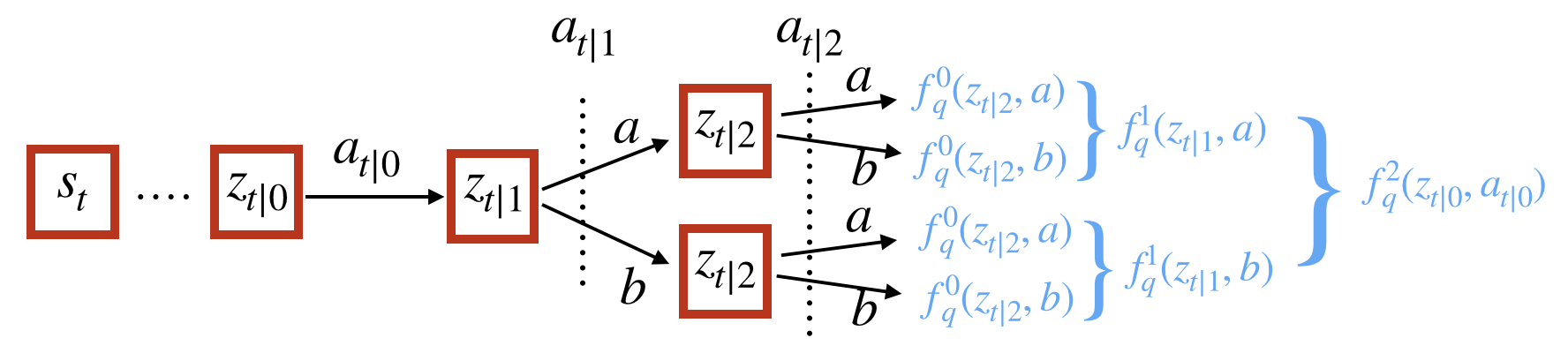}
\caption{\label{fig:tree} An example of the tree search (Equation~\ref{eq:tree-dpg}) with $d=2$ and $N=2$. We use $f_q^2(z_{t|0}, a_{t|0})$ as the estimation $Q(s_t, a_t)$. Here, $z_{t|l} (l=1,2)$ represents the predicted latent state after $l$ steps from $t$, $a$ and $b$ are actions proposed by the two actors and depend on latent state. Arrows represent the transition kernel $f_{trans}$, and brackets represent maximization (backup) operations. Reward prediction is omitted for simplicity.}
\end{figure*}

We use $f_{\mu_i}(z_{t|0})$ to represent $\mu_i(s_t)$ and $f_q(z_{t|0}, a_{t|0})$ to represent $Q(s_t, a_t)$, where $z_{t|0} = f_{enc}(s_t)$ is a latent state and $a_{t|0} = a_t$. Furthermore, $f_q(z_{t|0}, a_{t|0})$ can also be decomposed into the sum of the predicted immediate reward and the value of the predicted next latent state, i.e.,  
\begin{align}
f_q(z_{t|0}, a_{t|0}) \leftarrow f_{rew}(z_{t|0}, a_{t|0}) + \gamma f_q(z_{t|1}, a_{t|1})
\label{eq:fq}
\end{align}
where
\begin{align}
\label{eq:fq_1}
z_{t|1} &= f_{trans}(z_{t|0}, a_{t|0}) \\
\label{eq:fq_2}
a_{t|1} &= \text{argmax}_{a \in \{f_{\mu_i}(z_{t|1})\}_{i=1, \dots, N}} f_q({z_{t|1}, a})
\end{align}
We apply Equation~\eqref{eq:fq} recursively $d$ times with state prediction and action selection in Equations (\ref{eq:fq_1}, \ref{eq:fq_2}), resulting in a new estimator $f_q^d(z_{t|0}, a_{t|0})$ for $Q(s_t, a_t)$, which is defined as
\begin{align}
\label{eq:tree-dpg}
f_q^d(z_{t|l}, a_{t|l}) &= \begin{cases}
f_q(z_{t|l}, a_{t|l}) & d = 0 \\
f_{rew}(z_{t|l}, a_{t|l}) + \gamma f_{q}^{d-1}(z_{t|l+1}, a_{t|l+1})  & d > 0
\end{cases}
\end{align}
where 
\begin{align*}
z_{t|0} &\doteq f_{enc}(s_t), \quad a_{t|0} \doteq a_t \\
z_{t|l} &\doteq f_{trans}(z_{t|l-1}, a_{t|l-1}) \quad (l \geq 1) \\
a_{t|l} &\doteq \text{argmax}_{a \in \{ f_{\mu_i}(z_{t|l})\}_{i=1,\dots, N}} f_q^{d-l}(z_{t|l}, a) \quad (l \geq 1) 
\end{align*}
The look-ahead tree search and backup process (Equation~\ref{eq:tree-dpg}) are illustrated in Figure~\ref{fig:tree}. The value of $f_q^d(z_{t|l}, a_{t|l})$ stands for the state-action value estimation for the predicted latent state and action after $l$ steps from $t$, with Equation~\eqref{eq:fq} applied $d$ times.

As $f_q^d$ is fully differentiable w.r.t. $\theta^Q$, we plug in $f_q^d$ whenever we need $Q$. We also ground the predicted reward in the first recursive expansion as suggested by \citet{farquhar2018treeqn}. To summarize, given a transition $(s_t, a_t, r_{t+1}, s_{t+1})$, the gradients for $\theta^Q$ and $\theta^\mu$ are  
\begin{align*}
\nabla_{\theta^Q} \Bigg(&\frac{1}{2} \big(f_q^d(z_{t|0}, a_{t|0}) - \max_i f_q^d(z_{t+1|0}, f_{\mu_i}(z_{t+1|0}))\big)^2 \\
&+ \frac{1}{2}\big(f_{rew}(z_{t|0}, a_{t|0}) - r_{t+1}\big)^2 \Bigg)
\end{align*}
and
\begin{align*}
\sum_{i=1}^N \nabla_a f_q^d(z_{t|0}, a)|_{a=f_{\mu_i}(z_{t|0})} \nabla_{\theta^\mu} f_{\mu_i}(z_{t|0}) 
\end{align*}
respectively. ACE also utilizes experience replay and a target network similar to DDPG. The pseudo-code of ACE is provided in Supplementary Material.


\section{Experiments}
We designed experiments to answer the following questions:
\begin{itemize}
\item Does ACE outperform baseline algorithms?
\item If so, how do the components of ACE contribute to the performance boost?
\end{itemize}

We used twelve continuous control tasks from Roboschool, a free port of Mujoco~\footnote{http://www.mujoco.org/} by OpenAI. A state in Roboschool contains joint information of a robot (e.g., speed or angle) and is presented as a vector of real numbers. An action in Roboschool is a vector, with each dimension being a real number in $[-1, 1]$. All the implementations are made publicly available.~\footnote{https://github.com/ShangtongZhang/DeepRL}

\subsection{ACE Architecture}
In this section we describe the parameterization of $f_{enc}, f_{rew}, f_{trans}, f_q$ and $\{f_{\mu_i}\}_{i=1,\dots, N}$ for Roboschool tasks. For each state $s$, we first transformed it into a latent state $z \in \mathbb{R}^{400}$ by $f_{enc}$, which was parameterized as a single neural network layer. This latent state $z$ was used as the input for all other functions (i.e., $f_{rew}, f_q, f_{trans}, f_{\mu_1}, \dots, f_{\mu_N}$). 

\begin{figure*}[ht]
\centering
\includegraphics[width=0.8\textwidth]{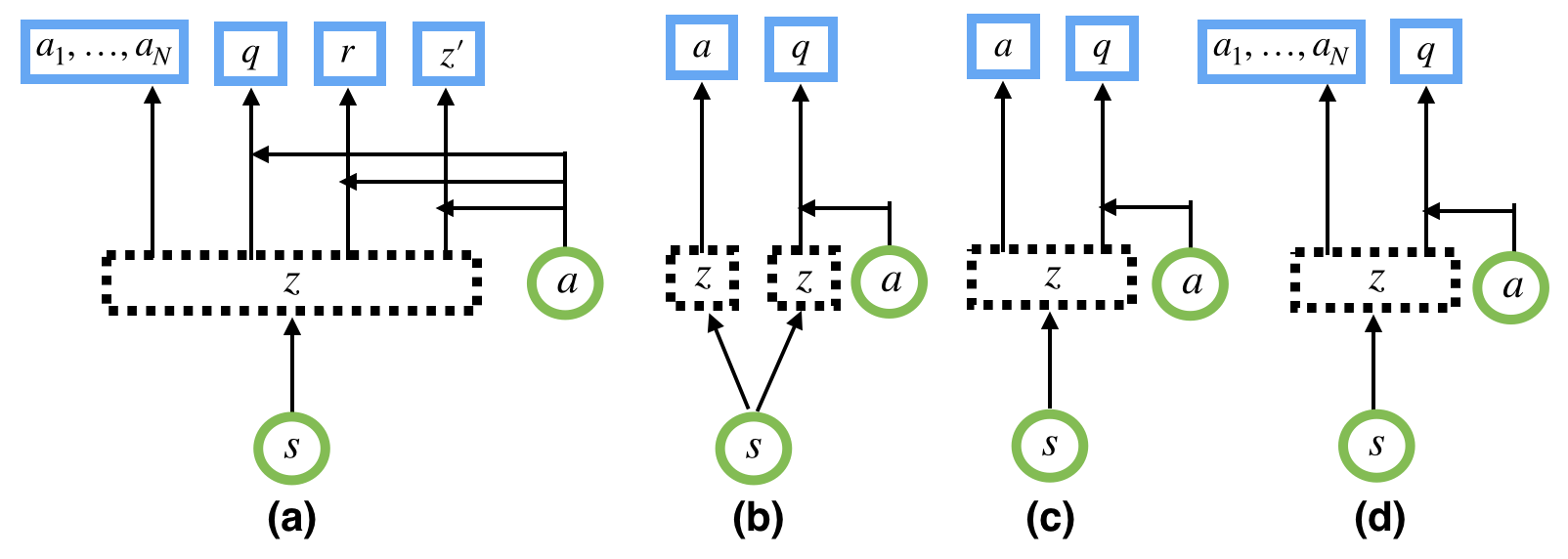}
\caption{\label{fig:arc} Architectures of the compared algorithms. (a) ACE and TM-ACE (b) DDPG and Wide-DDPG (c) Shared-DDPG (d) Ensemble-DDPG. Circles are input, dashed boxes are latent states, and solid boxes are outputs. A boxed $a_i$ is an action proposed by the $i$-th actor, while a circled $a$ is the actual action that the agent performed (which may include exploration noise). The shared layer of $a_1, \dots, a_N$ is not displayed for simplicity. }
\end{figure*}

The networks for $f_{rew}, f_q, f_{trans}$ are single hidden layer networks with $400+m$ input units and 300 hidden units, taking as inputs the concatenation of a latent state and an $m$-dimensional action. Particularly, the network for $f_{trans}$ used two residual connections similar to \citet{farquhar2018treeqn}. The networks for $\{f_{\mu_i}\}_{i=1, \dots, N}$ are single hidden layer networks with 400 input units and 300 hidden units, and all the $N$ networks of $\{f_{\mu_i}\}$ shared a common first layer. The architecture of ACE is illustrated in Figure~\ref{fig:arc}(a).

We used $\tanh$ as the activation functions for the hidden units. (This selection will be further discussed in the next section.) We set the number of actors to 5 (i.e., $N=5$) and the planning depth to 1 (i.e. $d=1$).

\subsection{Baseline Algorithms}

\subsubsection{DDPG} In DDPG, \citet{lillicrap2015continuous} used two separate networks to parameterize the actor and the critic. Each network had 2 hidden layers with 400 and 300 hidden units respectively. \citet{lillicrap2015continuous} used ReLU activation function (\citealt{nair2010rectified}) and applied a $L_2$ regularization to the critic. However, our analysis experiments found that $\tanh$ activation function outperformed ReLU with $L_2$ regularization. So throughout all our experiments, we always used $\tanh$ activation function (without $L_2$ regularization) for all algorithms. All other hyper-parameter values were the same as \citet{lillicrap2015continuous}. All the other compared algorithms inherited the hyper-parameter values from DDPG without tuning. We used the same Ornstein-Uhlenbeck process (\citealt{uhlenbeck1930theory}) as \citet{lillicrap2015continuous} for exploration in all the compared algorithms.

\subsubsection{Wide-DDPG} ACE had more parameters than DDPG. To investigate the influence of the number of parameters, we implemented Wide-DDPG, where the hidden units were doubled (i.e., the two hidden layers had 800 and 600 units respectively). Wide-DDPG had a comparable number of parameters as ACE and remained the same depth as ACE.

\subsubsection{Shared-DDPG} DDPG used separate networks for actor and critic, while the actor and critic in ACE shared a common representation layer. To investigate the influence of a shared representation, we implemented Shared-DDPG, where the actor and the critic shared a common bottom layer in the same manner as ACE.

\subsubsection{Ensemble-DDPG} To investigate the influence of the tree search in ACE, we removed the tree search in ACE by setting $d=0$, giving an ensemble version of DDPG, called Ensemble-DDPG. We still used 5 actors in Ensemble-DDPG. 

\subsubsection{TM-ACE} To investigate the usefulness of the value prediction model, we implemented Transition-Model-ACE (TM-ACE), where we learn a transition model instead of a value prediction model. To be more specific, $f_{trans}$ and $f_{rew}$ were trained to fit sampled transitions from the replay buffer to minimize the squared loss of the predicted reward and the predicted next latent state. This model was then used for a look-ahead tree search. The pseudo-code of TM-ACE is detailed in Supplementary Material.

The architectures of all the above algorithms are illustrated in Supplementary Material.

\subsection{Results}
For each task, we trained each algorithm for 1 million steps. At every 10 thousand steps, we performed 20 deterministic evaluation episodes without exploration noise and computed the mean episode return. We report the best evaluation performance during training in Table~\ref{tab:best}, which is averaged over 5 independent runs. The full evaluation curves are reported in Supplementary Material.

In a summary, either ACE or ACE-Alt was placed among the best algorithms in 11 out of the 12 games. ACE itself was placed among the best algorithms in 8 games taking ACE-Alt into comparison. Without considering ACE-Alt, ACE was placed among the best algorithms in 10 games. ACE-Alt itself was placed among the best algorithms in 7 games taking ACE into comparison. Without considering ACE, ACE-Alt was placed among the best algorithms in 10 games. Overall, ACE was slightly better than ACE-Alt. However, ACE-Alt enjoyed lower variance than ACE. We conjecture this is because ACE had more off-policy learning than ACE-Alt. Off-policy learning improved data efficiency but increased variances.

Wide-DDPG outperformed DDPG in only 1 game, indicating that naively increasing the parameters does not guarantee performance improvement. Shared-DDPG outperformed DDPG in only 2 games (lost in 2 games and tied in 8 games), showing shared representation contributes little to the overall performance in ACE. Ensemble-DDPG outperformed DDPG in 6 games (lost in 3 games and tied in 3 games), indicating the DDPG agent benefits from an actor ensemble. This may be attributed to that multiple actors are more likely to find the global maxima of the critic. ACE further outperformed Ensemble-DDPG in 9 games, indicating the agent benefits from the look-ahead tree search with a value prediction model. In contrast, TM-ACE outperformed Ensemble-DDPG only in 2 games (lost in 3 games and tied in 7 games), indicating that a value prediction model is better than a transition model for a look-ahead tree search. This was also consistent with the results observed earlier in VPN and TreeQN. 

In conclusion, the actor ensemble and the look-ahead tree search with a learned value prediction model are key to the performance boost. 

ACE and ACE-Alt increase performance in terms of environment steps while require more computation than vanilla DDPG. We benchmarked the wall time for the algorithms. The results are reported in Supplementary Material. We also verified the diversity of the actors in ACE and ACE-Alt in Supplementary Material. 

\subsection{Ensemble Size and Planning Depth}
\begin{figure}[h]
\centering
\includegraphics[width=0.4\textwidth]{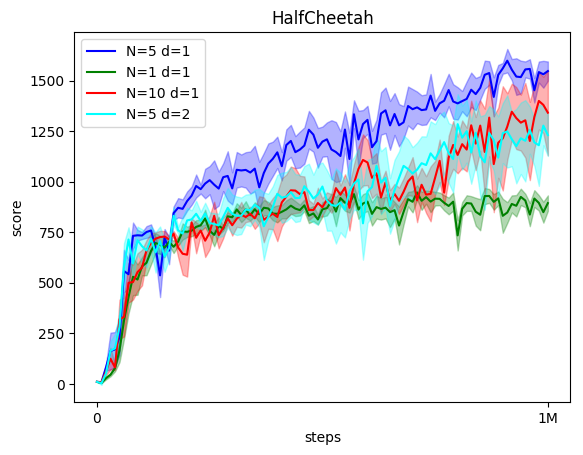}
\caption{\label{fig:half_cheetah} Evaluation curves of ACE in HalfCheetah with different $N$ and $d$. Each curve is averaged over 5 independent runs, and standard errors are plotted in shadow.}
\end{figure}

In this section, we investigate how the ensemble size $N$ and the planning depth $d$ in ACE influence the performance. We performed experiments in HalfCheetah with various $N$ and $d$ and used the same evaluation protocol as before. As a large $N$ and $d$ induced a significant computation increase, we only used $N$ up to 10 and $d$ up to 2. The results are reported in Figure~\ref{fig:half_cheetah}. 

To summarize, $N=5$ and $d=1$ achieved the best performance. We hypothesize there is a trade-off in the selection of both the ensemble size and the planning depth. On the one hand, a single actor can easily be trapped into local maxima during training. The more actors we have, the more likely we find the global maxima. On the other hand, all the actors share the same encoding function with the critic. A large number of actors may dominate the training of the encoding function to damage the critic learning. So a medium ensemble size is likely to achieve the best performance. A possible solution is to normalize the gradient according to the ensemble size, and we leave this for future work. With a perfect model, the more planning steps we have, the more accurate $Q$ estimation we can get. However, with a learned value prediction model, there is a compound error in unrolling. So a medium planning depth is likely to achieve the best performance. Similar phenomena were also observed in the multi-step Dyna planning (\citealt{yao2009multi}).

\begin{table*}[h]
\begin{center}
\small
\begin{tabular}{ |c|c|c|c|c|c|c|c| } 
\hline
 & ACE & ACE-Alt & TM-ACE & Ensemble-DDPG & Shared-DDPG & Wide-DDPG & DDPG \\
\hline
Ant &\textbf{1041}(70.8) &\textbf{983}(36.8) &\textbf{1031}(55.6) &\textbf{1026}(87.2) &796(16.8) &871(19.9) &875(14.2) \\
HalfCheetah &\textbf{1667}(40.4) &1023(60.4) &800(28.8) &812(49.4) &771(79.6) &733(52.5) &703(37.3) \\
Hopper &\textbf{2136}(86.4) &1923(88.3) &1586(85.0) &1972(63.4) &\textbf{2047}(76.7) &\textbf{2090}(118.6) &\textbf{2133}(99.0) \\
Humanoid &\textbf{380}(56.1) &\textbf{441}(90.1) &61(6.8) &76(11.4) &53(2.2) &54(1.0) &54(1.7) \\
HF &\textbf{311}(30.3) &\textbf{289}(20.8) &126(39.6) &85(5.6) &53(1.0) &55(1.6) &53(1.4) \\
HFH &\textbf{22}(2.4) &\textbf{20}(2.2) &-4(2.1) &2(7.5) &6(6.4) &15(3.2) &15(2.5) \\
IDP &7555(1610.9) &\textbf{9356}(1.1) &7549(1613.7) &4102(1923.2) &7549(1613.6) &7548(1618.7) &5662(1945.9) \\
IP &417(212.8) &\textbf{1000}(0.0) &415(213.4) &\textbf{1000}(0.0) &\textbf{1000}(0.0) &\textbf{1000}(0.0) &\textbf{1000}(0.0) \\
IPS &\textbf{892}(0.1) &892(0.2) &891(0.2) &891(0.2) &891(0.4) &546(308.7) &891(0.4) \\
Pong &\textbf{12}(0.3) &11(0.1) &6(0.7) &8(0.9) &4(0.2) &4(0.1) &5(0.8) \\
Reacher &16(0.7) &17(0.2) &17(0.5) &17(0.3) &\textbf{20}(0.7) &15(2.2) &18(0.9) \\
Walker2d &1659(65.9) &\textbf{1864}(21.4) &1086(97.0) &1142(99.5) &1142(146.3) &1185(121.2) &815(11.4) \\
\hline
\end{tabular}
\end{center}
\caption{\label{tab:best} Best evaluation performance during training. Mean and standard error are reported. Bold numbers indicate the best performance. Scores are averaged over 5 independent runs. Numbers are rounded for the ease of display. HF, HFH, IDP, IP, and IPS stand for HumanoidFlagrun, HumanoidFlagrunHarder, InvertedDoublePendulum, InvertedPendulum, and InvertedPendulumSwingup respectively.}
\end{table*}

\section{Related Work}

\subsubsection{Continuous-action RL} \citet{klissarov2017learnings} extended OCA into continuous action problems with the Proximal Policy Option Critic (PPOC) algorithm. However, PPOC considered stochastic intra-option policies, and each intra-option policy was trained via a policy search method. In ACE, we consider deterministic intra-option policies, and the intra-option policies are optimized under the same objective as OCA. 

\citet{gu2016continuous} parameterized the $Q$ function in a quadric form to deal with continuous control problems. In this way, the global maxima can be determined analytically. However, in general, the optimal $Q$ value does not necessarily fall into this quadric form. In ACE, we use an actor ensemble to search the global maxima of the $Q$ function. \citet{gu2016continuous} utilized a transition model to generate imaginary data, which is orthogonal to ACE.

\subsubsection{Ensemble in RL} \citet{wiering2008ensemble} designed four ensemble methods combining five RL algorithms with a voting scheme based on value functions of different RL algorithms. \citet{hans2010ensembles} used a network ensemble to improve the performance of Fitted Q-Iteration. \citet{osband2016deep} used a $Q$ ensemble to approximate Thomas' sampling, resulting in improved exploration and performance boost in challenging video games. \citet{huang2017learning} used both an actor ensemble and a critic ensemble in continuous control problems. However, to our best knowledge, the present work is the first to relate ensemble with options and to use an ensemble for a look-ahead tree search in continuous control problems. 

\section{Closing Remarks}
In this paper, we propose the ACE algorithm for continuous control problems. From an ensemble perspective, ACE utilizes an actor ensemble to search the global maxima of a critic function. From an option perspective, ACE is a special option-critic algorithm with deterministic intra-option policies. Thanks to the actor ensemble, ACE is able to perform a look-ahead tree search with a learned value prediction model in continuous control problems, resulting in a significant performance boost in challenging robot manipulation tasks.

\section{Acknowledgement}
The authors thank Bo Liu for feedbacks of the first draft of this paper. We also appreciate the insightful reviews of the anonymous reviewers.

{\bibliography{../ref.bib}}
\bibliographystyle{aaai}

\onecolumn
\section{Supplementary Material}
\subsection{Proof of Theorem~\ref{the:DIPG}}
Under mild conditions (\citealt{bacon2017option}), the Markov chain underlying $\pi_\Omega$ is aperiodic and ergodic. We use the following augmented process defined by \citet{bacon2017option}, which is homogeneous.
\begin{align*}
P^{(1)}_\gamma(s_{t+1}, \omega_{t+1}|s_t, \omega_t) &= \gamma p(s_{t+1}|s_t, \mu_{\omega_t}(s_t))\big(
(1 - \beta_{\omega_t}(s_{t+1}))\mathbb{I}_{\omega_t = \omega_{t+1}} + \beta_{\omega_t}(s_{t+1})\pi_\Omega(w_{t+1} | s_{t+1})\big) \\
P^{(k)}_\gamma(s_{t+k}, \omega_{t+k}|s_t, \omega_t) &= \sum_{s_{t+1}} 
\sum_{\omega_{t+1}}P^{(1)}_\gamma (s_{t+1}, \omega_{t+1}|s_t, \omega_t) P^{(k-1)}_\gamma (s_{t+k}, \omega_{t+k}|s_{t+1}, \omega_{t+1})
\end{align*}
We compute the gradient as
\begin{align}
\nabla_\theta Q_\Omega(s, \omega) &= \nabla_\theta Q_U(s, \omega, \mu_\omega(s)) \quad \text{(Equation~\ref{eq:three_q})} \nonumber \\ 
&=\nabla_\theta r(s, \mu_\omega(s)) + \gamma \nabla_\theta\sum_{s^\prime}p(s^\prime | s, \mu_\omega(s)) U(\omega, s^\prime) \quad \text{(Equation~\ref{eq:def_q_U})} \nonumber \\ 
&= \nabla_\theta \mu_\omega(s) \nabla_a r(s, a)_{|a=\mu_\omega(s)} + \gamma \sum_{s^\prime}\nabla_\theta p(s^\prime | s, \mu_\omega(s)) U(\omega, s^\prime) + \gamma \sum_{s^\prime} p(s^\prime | s, \mu_\omega(s)) \nabla_\theta U(\omega, s^\prime) \nonumber \\
&= \nabla_\theta \mu_\omega(s) \nabla_a r(s, a)_{|a=\mu_\omega(s)} \nonumber + \gamma \sum_{s^\prime}\nabla_\theta \mu_\omega(s) \nabla_a p(s^\prime | s, a)_{|a=\mu_\omega(s)} U(\omega, s^\prime) + \gamma \sum_{s^\prime} p(s^\prime | s, \mu_\omega(s)) \nabla_\theta U(\omega, s^\prime) \nonumber \\
&= \nabla_\theta \mu_\omega(s) \nabla_a \big( r(s, a) + \gamma \sum_{s^\prime}p(s^\prime | s, a)U(\omega, s^\prime)\big)_{|a=\mu_\omega(s)} + \gamma \sum_{s^\prime} p(s^\prime | s, \mu_\omega(s)) \nabla_\theta U(\omega, s^\prime) \nonumber \\
\label{eq:dQ-part1}
&= \nabla_\theta \mu_\omega(s) \nabla_a Q_U(s, \omega, a)_{|a=\mu_\omega(s)} + \gamma \sum_{s^\prime} p(s^\prime | s, \mu_\omega(s)) \nabla_\theta U(\omega, s^\prime) \quad \text{(Equation~\ref{eq:def_q_U})} 
\end{align}
From Equation~\eqref{eq:def_U}, we have 
\begin{align}
\label{eq:dQ-part2}
\nabla_\theta U(\omega, s^\prime) &= (1 - \beta_\omega(s^\prime))\nabla_\theta Q_\Omega(s^\prime, \omega) + \beta_\omega(s^\prime)\nabla_\theta V_\Omega(s^\prime) \nonumber \\
&= (1 - \beta_\omega(s^\prime))\nabla_\theta Q_\Omega(s^\prime, \omega) + \beta_\omega(s^\prime) \sum_{\omega^\prime} \pi_\Omega(\omega^\prime | s^\prime) \nabla_\theta Q_\Omega(s^\prime, \omega^\prime) \nonumber \\
&= \sum_{\omega^\prime} \big( (1 - \beta_\omega(s^\prime))\mathbb{I}_{\omega=\omega^\prime} + \beta_\omega(s^\prime) \pi_\Omega(\omega^\prime | s^\prime)\big) \nabla Q_\Omega(s^\prime, \omega^\prime)
\end{align}
Plug in Equation~\eqref{eq:dQ-part2} into Equation~\eqref{eq:dQ-part1} and use the definition of $P^{(1)}_\gamma$, we have
\begin{align}
\label{eq:recursive_dQ}
\nabla_\theta Q_\Omega(s, \omega) &= \nabla_\theta \mu_\omega(s) \nabla_a Q_U(s, \omega, a)_{|a=\mu_\omega(s)} + \sum_{s^\prime, \omega^\prime} P^{(1)}_\gamma (s^\prime, \omega^\prime | s, \omega) \nabla_\theta Q_\Omega(s^\prime, \omega^\prime)
\end{align}
Expand $\nabla_\theta Q_\Omega(s_0, \omega_0)$ with Equation~\eqref{eq:recursive_dQ} recursively and apply the augmented process, we end up with 
\begin{align*}
\nabla_\theta Q_\Omega(s_0, \omega_0) = \sum_{s, \omega}\rho_\Omega(s, \omega)\nabla_a Q_U(s, \omega, a)|_{a=\mu_\omega(s)}\nabla_\theta \mu_\omega(s)
\end{align*}

\subsection{Proof of Theorem~\ref{the:TPG}}
Under mild conditions (\citealt{bacon2017option}), the Markov chain underlying $\pi_\Omega$ is aperiodic and ergodic. We use the following augmented process defined by \citet{bacon2017option}, which is homogeneous. 
\begin{align*}
P^{(1)}_\gamma(s_{t+1}, \omega_{t}|s_t, \omega_{t-1}) &= \gamma p(s_{t+1}|s_t, \mu_{\omega_{t-1}}(s_t))\big(
(1 - \beta_{\omega_{t-1}}(s_{t+1}))\mathbb{I}_{\omega_{t-1} = \omega_{t}} + \beta_{\omega_{t-1}}(s_{t+1})\pi_\Omega(w_{t} | s_{t+1})\big)
\end{align*}
\begin{align*}
P^{(k)}_\gamma(s_{t+k}, \omega_{t+k-1}|s_t, \omega_{t-1}) &= \sum_{s_{t+1}} 
\sum_{\omega_{t}}P^{(1)}_\gamma (s_{t+1}, \omega_{t}|s_t, \omega_{t-1}) P^{(k-1)}_\gamma (s_{t+k}, \omega_{t+k-1}|s_{t+1}, \omega_{t})
\end{align*}
We have 
\begin{align}
\label{eq:dQ_dnu}
\nabla_\nu Q_\Omega(s^\prime, \omega) &= \nabla_\nu Q_\Omega(s^\prime, \omega, \mu_\omega(s^\prime)) \nonumber \\
&= \nabla_\nu \Big(r(s^\prime, \mu_\omega(s^\prime)) + \gamma \sum_{s^{\prime\prime}} p(s^{\prime\prime}|s^\prime, \mu_\omega(s^\prime))U(\omega, s^{\prime\prime})\Big) \nonumber \quad (\text{Equation~\ref{eq:def_q_U}}) \\ 
&= \gamma \sum_{s^{\prime\prime}} p(s^{\prime\prime}|s^\prime, \mu_\omega(s^\prime)) \nabla_\nu U(\omega, s^{\prime\prime}) 
\end{align}
The gradient of $U(\omega, s^\prime)$ w.r.t. $\nu$ is
\begin{align}
\nabla_\nu U(\omega, s^\prime) &= Q_\Omega(s^\prime, \omega) \nabla_\nu (1 - \beta_\omega(s^\prime)) + (1 - \beta_\omega(s^\prime))\nabla_\nu Q_\Omega(s^\prime, \omega) \nonumber \\
& \quad + \beta_\omega(s^\prime) \nabla_\nu V_\Omega(s^\prime) + V_\Omega(s^\prime) \nabla_\nu \beta_\omega(s^\prime) \nonumber \quad \text{(Equation~\ref{eq:def_U} and product rule of calculus)} \\
&= Q_\Omega(s^\prime, \omega) \nabla_\nu (1 - \beta_\omega(s^\prime)) + (1 - \beta_\omega(s^\prime))\nabla_\nu Q_\Omega(s^\prime, \omega) \nonumber \\
& \quad + \beta_\omega(s^\prime) \nabla_\nu \sum_{\omega^\prime}\pi_\Omega(\omega^\prime | s^\prime) Q_\Omega(s^\prime, \omega^\prime) + V_\Omega(s^\prime) \nabla_\nu \beta_\omega(s^\prime) \nonumber \quad \text{(Equation~\ref{eq:def_V})}\\
&= \nabla_\nu \beta_\omega(s^\prime)(V_\Omega(s^\prime) - Q_\Omega(s^\prime, \omega)) \nonumber \\
&\quad + (1 - \beta_\omega(s^\prime))\nabla_\nu Q_\Omega(s^\prime, \omega) + \beta_\omega(s^\prime) \sum_{\omega^\prime}\pi_\Omega(\omega^\prime | s^\prime) \nabla_\nu Q_\Omega(s^\prime, \omega^\prime) \nonumber\\
%
&= \nabla_\nu \beta_\omega(s^\prime)(V_\Omega(s^\prime) - Q_\Omega(s^\prime, \omega)) \nonumber \\
& \quad + \sum_{s^{\prime\prime}} \sum_{\omega^\prime} \big( (1-\beta_\omega(s^\prime))\mathbb{I}_{\omega^\prime = \omega} + \beta_\omega(s^\prime)\pi_\Omega(\omega^\prime | s^\prime) \big)\gamma p(s^{\prime\prime}|s^\prime, \mu_{\omega^\prime}(s^\prime))\nabla_\nu U(\omega^\prime, s^{\prime\prime}) \nonumber \quad \text{(Equation~\ref{eq:dQ_dnu})} \\
\label{eq:du-part1}
&= \nabla_\nu \beta_\omega(s^\prime)(V_\Omega(s^\prime) - Q_\Omega(s^\prime, \omega)) + \sum_{s^{\prime\prime}} \sum_{\omega^\prime} P^{(1)}_\gamma (\omega^\prime, s^{\prime\prime} | \omega, s^\prime) \nabla_\nu U(\omega^\prime, s^{\prime\prime}) \quad \text{(Definition of $P^{(1)}_\gamma$)}
\end{align} 
Applying Equation~\eqref{eq:du-part1} recursively and using the augmented process, we have
\begin{align*}
\nabla_\nu U(\omega_0, s_1) &= \sum_{\omega, s^\prime}\rho_\Omega(s^\prime, \omega|s_1, \omega_0)\nabla_\nu \beta_\omega(s^\prime)(V_\Omega(s^\prime) - Q_\Omega(s^\prime, \omega))
\end{align*}

\subsection{The ACE Algorithm}
\begin{algorithm}[ht]
\DontPrintSemicolon
\textbf{Input:}\\
$N$: number of actors \\
$\epsilon_t$: a noise process \\ 
$d$: plan depth \\
$\alpha, \beta$: two step sizes \\
$f_q, f_{enc}, f_{rew}, f_{trans}, f_{\mu_1}, \dots, f_{\mu_N}$: parameterized by $\theta^Q$ and $\theta^\mu$ \quad \tcp{\small see Section Model-based Enhancements}
\textbf{Output:}\\
The parameters $\theta^Q$ and $\theta^\mu$ \\
\hrulefill \\
Initialize the replay buffer $\mathcal{D}$ \\
\For{each time step t} {
	Observe the state $s_t$\\
	$z_t \leftarrow f_{enc}(s_t)$ \\
	$a_t \leftarrow \text{argmax}_{a \in \{f_{\mu_i}(z_t)\}_{i=1,\dots, N}} f_q^d(z_t, a) + \epsilon_t$ \quad \tcp{\small see Equation~\ref{eq:tree-dpg} for $f_q^d$} 
	Execute the action $a_t$, get reward $r_{t+1}$ and next state $s_{t+1}$ \\ 
	Store $(s_t, a_t, r_{t+1}, s_{t+1})$ into $\mathcal{D}$ \\
	Sample a batch of transitions $\mathcal{B}$ from $\mathcal{D}$ \\
	\For{each transition $s, a, r, s^\prime$ in $\mathcal{B}$} {
		$z \leftarrow f_{enc}(s)$ \\
		$z^\prime \leftarrow f_{enc}(s^\prime)$ \\ 
		$y \leftarrow \begin{cases} 0  & \text{if $s^\prime$ is terminal}\\
		\max_i f_q^d(z^\prime, f_{\mu_i}(z^\prime)) & \text{otherwise}
		\end{cases}$ \\
		$y \leftarrow r + \gamma y$ \\
		$\hat{r} \leftarrow f_{rew}(z, a)$ \\
		$\theta^Q \leftarrow \theta^Q - \alpha \nabla_{\theta^Q} \Big(\frac{1}{2}(f_q^d(z, a) - y)^2 + \frac{1}{2}(\hat{r} - r)^2$ \Big)\\
		$\theta^\mu \leftarrow \theta^\mu + \beta \nabla_{\theta^\mu} \sum_{i=1}^N \nabla_b f_q^d(z, b)|_{b=f_{\mu_i}(z)} \nabla_{\theta^\mu} f_{\mu_i}(z) $ \\
	}
}
\caption{\label{alg:tree-dpg}ACE}
\end{algorithm}

\newpage
\subsection{ACE with a transition model}
TM-ACE only performs a look-ahead tree search in decision making. During training, TM-ACE uses only $f_q$ instead of $f_q^d$. 
\begin{algorithm}[ht]
\DontPrintSemicolon
\textbf{Input:}\\
$N$: number of actors \\
$\epsilon_t$: a noise process \\ 
$d$: plan depth \\
$\alpha, \beta$: two step size \\
$f_q, f_{enc}, f_{rew}, f_{trans}, f_{\mu_1}, \dots, f_{\mu_N}$: parameterized by $\theta^Q$ and $\theta^\mu$ \quad \tcp{\small see Section Model-based Enhancements}
\textbf{Output:}\\
The parameters $\theta^Q$ and $\theta^\mu$ \\
\hrulefill\\
Initialize the replay buffer $\mathcal{D}$ \\
\For{each time step t} {
	Observe the state $s_t$ \\
	$z_t \leftarrow f_{enc}(s_t)$ \\
	$a_t \leftarrow \text{argmax}_{a \in \{f_{\mu_i}(z_t)\}_{i=1,\dots, N}} f_q^d(z_t, a) + \epsilon_t$ \quad \tcp{\small see Equation~\ref{eq:tree-dpg} for $f_q^d$}
	Execute the action $a_t$, get reward $r_{t+1}$ and next state $s_{t+1}$ \\ 
	Store $(s_t, a_t, r_{t+1}, s_{t+1})$ into $\mathcal{D}$ \\
	Sample a batch of transitions $\mathcal{B}$ from $\mathcal{D}$ \\
	\For{each transition $s, a, r, s^\prime$ in $\mathcal{B}$} {
		$z \leftarrow f_{enc}(s)$ \\
		$z^\prime \leftarrow f_{enc}(s^\prime)$ \\ 
		$y \leftarrow \begin{cases} 0  & \text{if $s^\prime$ is terminal}\\
		\max_i f_q(z^\prime, f_{\mu_i}(z^\prime)) & \text{otherwise}
		\end{cases}$ \\
		$y \leftarrow r + \gamma y$ \\
		$\hat{r} \leftarrow f_{rew}(z, a)$ \\
		$\theta^Q \leftarrow \theta^Q - \alpha \nabla_{\theta^Q} \Big(\frac{1}{2}(f_q(z, a) - y)^2 + \frac{1}{2}(\hat{r} - r)^2 + \frac{1}{2}||f_{trans}(z, a) - z^\prime||^2$ \Big)\\
		$\theta^\mu \leftarrow \theta^\mu + \beta \nabla_{\theta^\mu} \sum_{i=1}^N \nabla_b f_q(z, b)|_{b=f_{\mu_i}(z)} \nabla_{\theta^\mu} f_{\mu_i}(z) $ \\
	}
}
\caption{TM-ACE}
\end{algorithm}

\subsection{Wall time comparison}
We benchmarked the wall time for ACE and DDPG with an i7-8750H CPU (2.20GHz).

\begin{table*}[h]
\begin{center}
\small
\begin{tabular}{ |c|c| } 
\hline
 & environment steps per second \\
\hline
DDPG & 73.63 \\
ACE with $d=1, N=5$ & 27.82 \\
ACE with $d=1, N=10$ & 16.12 \\
ACE with $d=2, N=5$ & 6.34 \\
\hline
\end{tabular}
\end{center}
\caption{Wall time for different algorithms in HalfCheetah}
\end{table*}

\newpage
\subsection{Evaluation curves}

\begin{figure*}[h]
\centering
\includegraphics[width=1.0\textwidth]{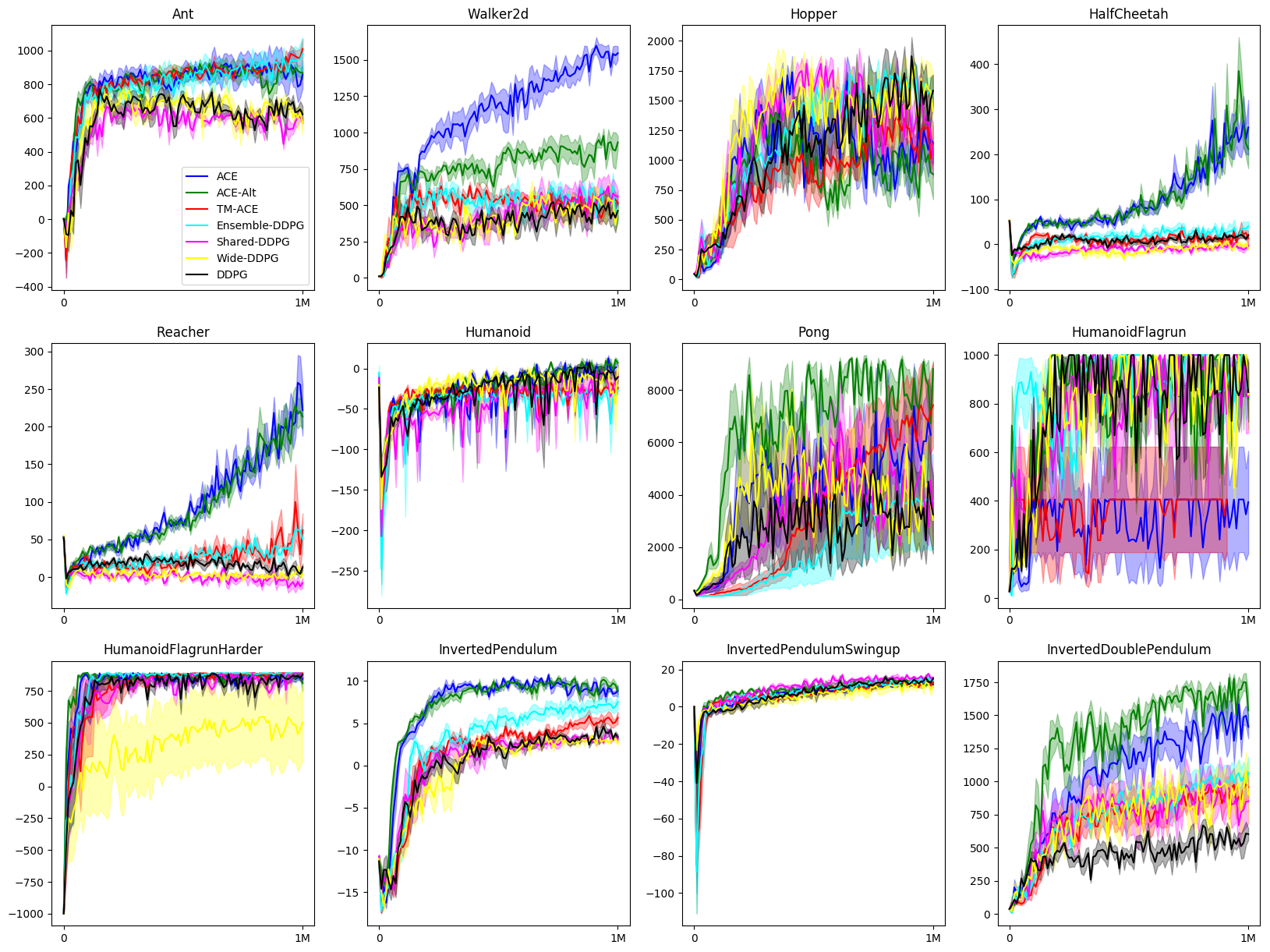}
\caption{\label{fig:curves} Evaluation curves of the compared algorithms in the Roboschool tasks. The x-axis is the training steps, and the y-axis is the raw evaluation scores. Each curve is averaged over 5 independent runs, and standard errors are plotted as shadow.}
\end{figure*}

\subsection{Diversity of the actors}
To verify the diversity of the actors, we evaluated each actor separately at the middle of training. We reported the average episode return of 20 deterministic episodes for each actor in ACE and ACE-Alt. In both ACE and ACE-Alt, the actors can roughly be grouped into four groups, indicating those actors are seeking for different policies. It is also observed that actors in ACE-Alt have more overall variance than actors in ACE, which is as expected as transitions are not shared among different actors in ACE-Alt.

\begin{table*}[h]
\begin{center}
\small
\begin{tabular}{ |c|c|c|c|c|c| } 
\hline
 & actor 1 & actor 2 & actor 3 & actor 4 & actor 5 \\
\hline
ACE & 714.8 (14.7) & 775.5 (4.9) & 550.8 (54.9) & 771.2 (5.0) & 645.7 (41.9) \\
ACE-Alt & 418.8 (30.1) & 631.3 (96.8) & 302.7 (44.0) & 945.2 (44.9) & 916.2 (16.0) \\
\hline
\end{tabular}
\end{center}
\caption{Evaluation performance of individual actors in ACE and ACE-Alt (with $d=2, N=5$) in HalfCheetah}
\end{table*}

\newpage
\subsection{Architectures of the algorithms}

\end{document}